\documentclass[preprint,12pt,authoryear]{elsarticle}
\usepackage{graphicx}
\usepackage{booktabs}
\usepackage{multirow}
\usepackage{amsmath}
\usepackage{amssymb}
\usepackage[caption=false, font=footnotesize]{subfig}
\usepackage[dvipsnames]{xcolor}
\usepackage{threeparttable}
\usepackage{hyperref}
\usepackage[normalem]{ulem}
\usepackage{xcolor}
\usepackage{algorithm}
\usepackage{algorithmic}
\newcommand{\firstbest}[1]{{\color{ForestGreen}\textbf{#1}}}
\newcommand{\secondbest}[1]{{\color{RoyalBlue}\textbf{#1}}}
\usepackage{url}
\usepackage{lineno}

\journal{International Journal of Applied Earth Observation and Geoinformation}

\begin{document}

\begin{frontmatter}

\title{Enhancing Monocular Height Estimation via Weak Supervision from Imperfect Labels}

\author[label1,label2]{Sining~Chen}
\author[label3]{Yilei~Shi}
\author[label1,label2]{Xiao~Xiang~Zhu} %% Author name
\affiliation[label1]{organization={Chair of Data Science in Earth Observation, Technical University of Munich (TUM)},
            city={Munich},
            postcode={80333},
            country={Germany}}

\affiliation[label2]{organization={Munich Center for Machine Learning (MCML)},
            city={Munich},
            postcode={80538},
            country={Germany}}
            
\affiliation[label3]{organization={School of Engineering and Design, Technical University of Munich (TUM)},
            city={Munich},
            postcode={80333},
            country={Germany}}
\begin{abstract}
Monocular height estimation provides an efficient and cost-effective solution for three-dimensional perception in remote sensing. However, training deep neural networks for this task demands abundant annotated data, while high-quality labels are scarce and typically available only in developed regions, which limits model generalization and constrains their applicability at large scales.
This work addresses the problem by leveraging imperfect labels from out-of-domain regions to train pixel-wise height estimation networks, which may be incomplete, inexact, or inaccurate compared to high-quality annotations.
We introduce an ensemble-based pipeline compatible with any monocular height estimation network, featuring architecture and loss functions specifically designed to leverage information in noisy labels through weak supervision, utilizing balanced soft losses and ordinal constraints.
Experiments on two datasets---DFC23 (0.5--1 m) and GBH (3 m)---show that our method achieves more consistent cross-domain performance, reducing average RMSE by up to 22.94\% on DFC23 and 18.62\% on GBH compared with baselines. Ablation studies confirm the contribution of each design component.
\end{abstract}

% \begin{highlights}
% \item First work to train monocular height estimation models with imperfect labels.
% \item Novel pipeline designed to leverage noisy and incomplete supervision.
% \item Ensemble network adapts to different label qualities effectively.
% \item Tailored loss functions optimize learning from imperfect annotations.
% \item Achieves up to 22.94\% improvement, especially in out-of-domain areas.
% \end{highlights}

%% Keywords

\end{frontmatter}
% \linenumbers
\section{Introduction}
Three-dimensional perception in remote sensing underpins numerous applications by providing essential spatial information. In particular, 3D modeling of buildings enhances our understanding of the built environment \cite{schug2021}, since building heights---and the building volumes derived from them---exhibit strong correlations with urban anthropogenic factors such as population density \cite{global3d}, energy consumption \cite{shareef2021}, solar potential \cite{li2024b}, and urban heat island effects \cite{wang2021g}. Among 3D perception methods, monocular height estimation---inferring height from single remote sensing images---is both efficient and cost-effective, making it suitable for large-scale applications, especially in resource-limited regions of the Global South.

Since the advent of deep learning, extensive research has been conducted on monocular height estimation using deep learning-based methods. Early studies employed generic fully convolutional neural networks \citep{ronneberger2015,badrinarayanan2017}, while later works introduced tailored architectures to enhance performance \citep{srivastava2017,mou2018,amirkolaee2019,xing2021,paoletti2021,chen2023e}. Despite these advances, current methods still depend on large quantities of well-annotated training data and rarely consider a critical real-world constraint: the scarcity of perfect labels. High-quality annotations typically come from Light Detection and Ranging (LiDAR) observations, which are costly and therefore concentrated in developed regions such as North America and Europe. Models trained exclusively on such data often generalize poorly to unseen domains---particularly in high-demand developing regions (e.g., South America and Asia)---where only imperfect labels derived from alternative sources exist.

If training continues to rely solely on LiDAR-derived labels, performance in these label-scarce regions will remain limited due to domain shifts in landscape, architectural style, and imaging conditions. While alternative strategies such as synthetic data generation and domain adaptation \citep{thebenchmark,xiong2022,zhao2023b,song2024} can alleviate some of these issues, they cannot fully replace real geospatial measurements. Incorporating imperfect labels provides a practical and scalable approach to extending model applicability worldwide, enabling reasonable performance in regions where perfect labels are unlikely to become available in the foreseeable future.

Several attempts have been made to train networks using imperfect data through weakly supervised learning \citep{chen2017,zhou2018a,lienen2021,xiong2022}, but most are task-specific and not directly applicable to monocular height estimation. Developing a framework specifically for this task requires addressing several factors. First, monocular height estimation involves varying levels of label quality, each necessitating distinct treatment due to different label properties. Second, the availability of high-quality labels depends on the regional development status, resulting in domain variation that is aligned with label quality. Third, a particularly challenging issue is the long-tailed distribution of height values, which is also present within imperfect labels \citep{chen2023e,chen2023a}. 

To address these challenges, we propose an ensemble-based framework capable of handling diverse levels of label quality. The training loss functions are carefully designed to mitigate the associated difficulties. Our main contributions are summarized as follows:
\begin{itemize}
    \item To the best of our knowledge, this work is the first to address the integration of imperfect labels in training monocular height estimation networks.
    \item We introduce an ensemble-based pipeline compatible with various label qualities.
    \item We develop loss functions tailored to leverage supervisory signals from imperfect labels, including a balanced soft height loss and ordinal constraints. Additionally, we propose a ground truth augmentation module to reduce the blurring effect in model outputs.
    \item We conduct extensive experiments on two datasets, demonstrating the effectiveness of our pipeline in learning from imperfect labels and achieving balanced performance across diverse domains.
\end{itemize}
\section{Related Works}\label{chap:related_works}
\subsection{Monocular Height Estimation}
Monocular height estimation predicts height values from single optical images. Early approaches explicitly utilized shadows as cues \citep{izadi2012}, but their reliance on visible shadows limits applicability in dense urban areas. With deep learning, methods based on neural networks have become dominant due to their flexibility and superior performance. 

Here, monocular height estimation denotes pixel-wise height prediction, where a height value is estimated for each pixel. This requires neural networks for dense predictions, e.g., fully convolutional neural networks (FCNs). General FCNs such as U-Net \citep{ronneberger2015} and SegNet \citep{badrinarayanan2017} can be adapted for this purpose by omitting the final activation layer. These encoder-decoder structures reconstruct outputs matching the input resolution, incorporating enhancements such as skip connections that fuse low- and high-level features.

Simultaneously, several networks have been specifically designed for monocular height estimation \citep{srivastava2017,mou2018,amirkolaee2019,xing2021,paoletti2021,chen2023e}. \citet{mou2018} proposed one of the pioneering lightweight encoder-decoder networks featuring a single skip connection for low-level feature fusion. \citet{amirkolaee2019} extended the U-Net architecture with upsampling blocks and multi-level feature fusion. \citet{xing2021} introduced PLNet, which features gated feature aggregation and progressive refinement modules. \citet{chen2023e} proposed HTC-DC Net, adopting a classification-regression paradigm inspired by \citet{bhat2021}, and enhanced via head-tail cut (HTC) and distribution-based constraints (DC). Multi-task learning \citep{srivastava2017} and generative adversarial training \citep{paoletti2021} have also been explored.

Despite significant progress, most studies assume abundant high-quality labels, overlook the label scarcity in developing regions, and rarely test generalization across diverse domains. In practice, this limitation significantly impacts the applicability of monocular height estimation models and warrants further investigation.

To address label scarcity, some studies have explored the generation of synthetic data and domain adaptation. \citet{thebenchmark} created a large-scale synthetic dataset using the video game GTA and transferred knowledge to real-world data via few-shot learning. Building upon this work, they further explored knowledge transfer in a label-efficient manner via weakly-supervised learning \citep{xiong2022}. \citet{song2024} synthesized a dataset for global 3D semantic understanding and proposed RS3DAda, a pipeline for domain adaptation in real-world settings. \citet{zhao2023b} proposed a semantic-aware domain adaptation approach combining image translation and multi-task training to enhance robustness. These strategies, however, mainly mitigate domain gaps between synthetic and real data or between different sensors, but not those caused by geographical and label-quality variations---issues central to global-scale height estimation.
\subsection{Weakly Supervised Learning}
Weakly supervised learning aims to train models with imperfect supervision, thereby reducing the intensive labor and high cost of obtaining large-scale, high-quality datasets \citep{zhou2018a}. It encompasses learning from incomplete, inexact, or inaccurate supervision \citep{zhou2018a}. In classification, for example, semi-supervised learning leverages unlabeled data as incomplete supervision \citep{sohn2020b,huang2023,heidler2024}. Multi-instance learning addresses inexact labels by assigning supervision to instance bags \citep{zhou2006,zhou2007}, while noisy-label learning corrects inaccurate supervision through data editing \citep{muhlenbach2004,liu2024a}.

Weakly supervised regression remains less studied. Different from classification, regression is challenging due to its continuous output space. A related example is monocular depth estimation in computer vision, where networks learn from relative rather than absolute depth. \citet{chen2017} trained models using ordinal relations---whether one pixel is nearer, farther, or equally distant---encoded via an ordinal loss. \citet{lienen2021} extended this approach by ranking pixel lists with a Plackett-Luce model. Inspired by these works, \citet{xiong2022} generated weak pseudo-labels from a synthetic model and fine-tuned them on real data for label-efficient monocular height estimation.

Although weakly supervised learning has been explored in monocular depth and height estimation, existing approaches do not adequately address the real-world issues of label quality and model generalizability. Consequently, they cannot be directly adapted to the problem addressed in this work, underscoring the need to develop novel frameworks.
\section{Methodology}\label{chap:methodology}
\subsection{Problem Formulation}
\begin{figure*}[!ht]
    \centering
    \includegraphics[width=\linewidth]{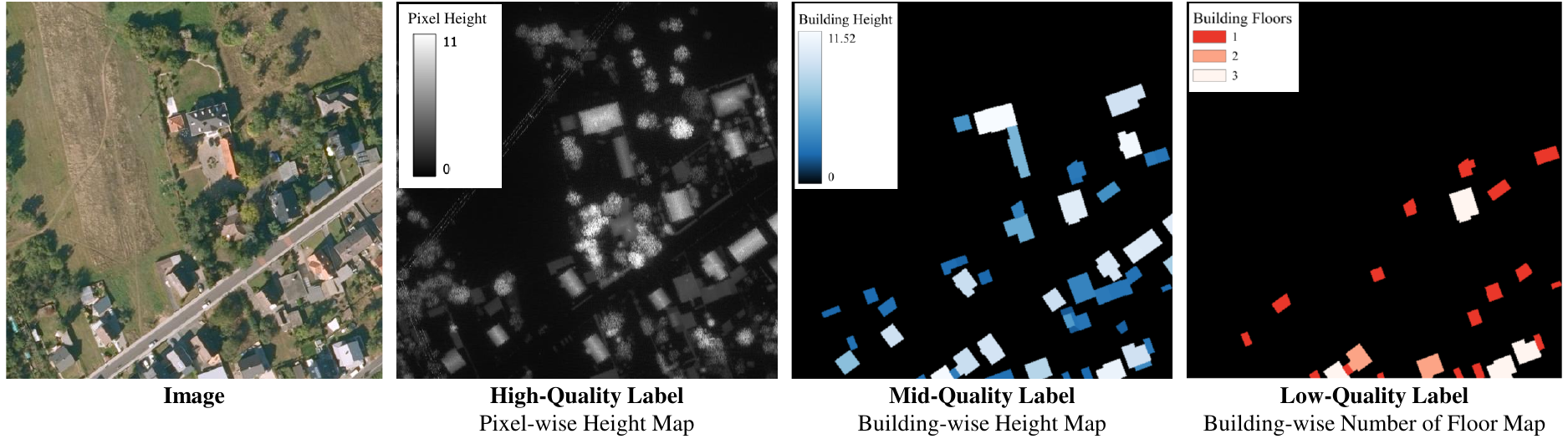}
    \caption{Common label qualities for monocular height estimation. Data from \citet{BayerischeVermessungsverwaltung_2025_OpenData}.}
    \label{fig:label_qualities}
\end{figure*}
The objective is to train a neural network for pixel-wise height estimation using labels of varying quality (Fig. \ref{fig:label_qualities}). These labels fall into three levels: high-, mid-, and low-quality.
\begin{itemize}
    \item \textbf{High-quality labels} closely match the target output, typically pixel-wise height maps obtained from LiDAR, delivered as normalized digital surface models (nDSMs). They provide accurate local structure but have limited spatial coverage due to acquisition costs. 
    \item \textbf{Mid-quality labels} provide instance-level (building-wise) heights derived from 3D building models, often simplified Level of Detail (LoD)-1 blocks assigning a single height per building. These are reliable at the building scale but lack fine structural details, resulting in oversmoothed pixel-wise supervision.
    \item \textbf{Low-quality labels} are noisier and less complete, generally representing building-wise floor counts. While correlated with physical height, variable floor heights introduce significant noise into the labels \citep{biljecki2017}. Nevertheless, such data are widely available from census or volunteered geographic sources such as OpenStreetMap \citep{OpenStreetMap}.
\end{itemize}
\subsection{Network Architecture}
\begin{figure*}[!t]
    \centering
    \includegraphics[width=\linewidth]{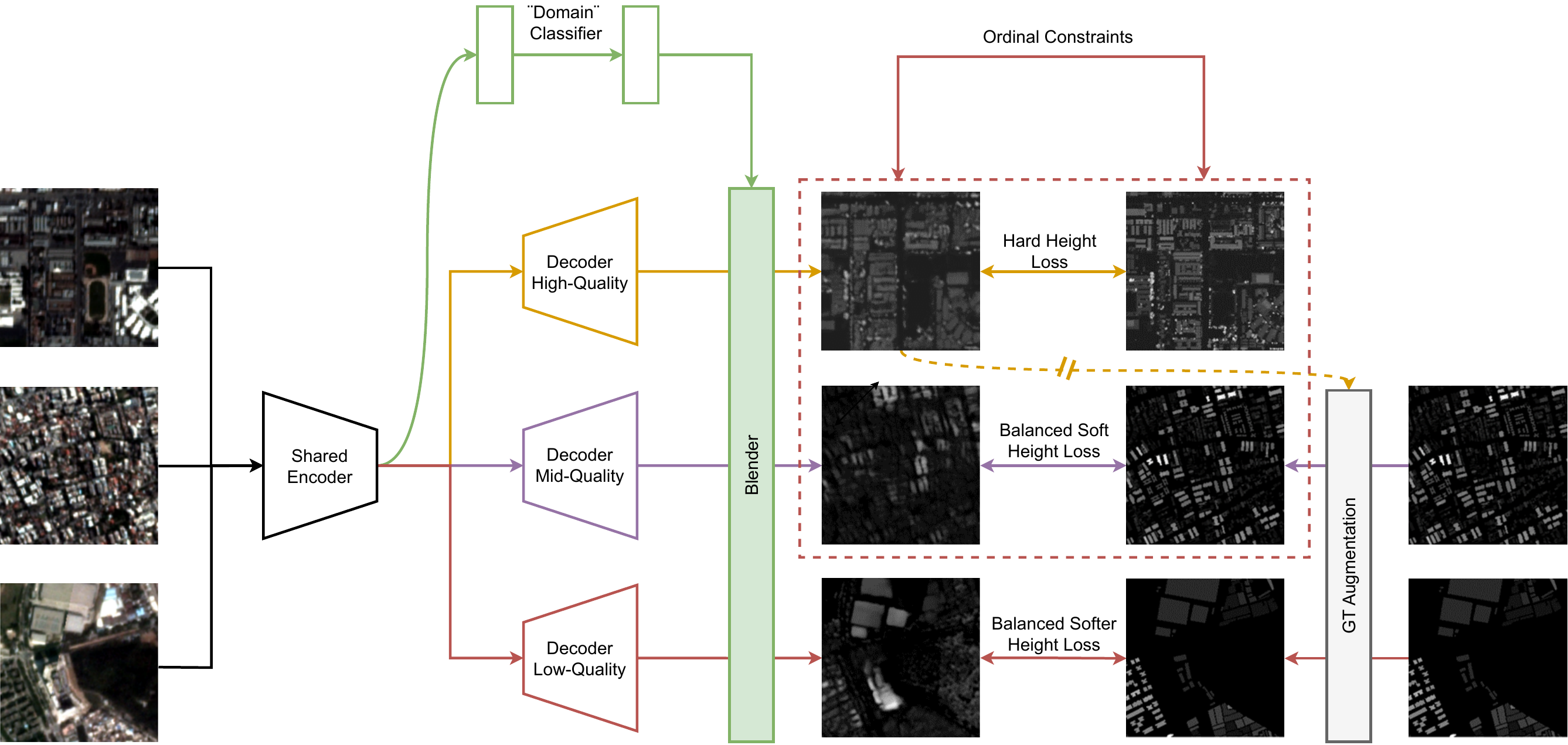}
    \caption{Network architecture. The proposed ensemble-based network consists of separate branches dedicated to high-, mid-, and low-quality labels. A "domain" classifier is employed to distinguish between label qualities and blend the outputs from each branch into the final prediction. Additionally, outputs from the high-quality branch are used for ground truth augmentation for the mid- and low-quality labels, with backward gradients stopped to prevent interference.}
    \label{fig:network}
\end{figure*}
As illustrated in Fig. \ref{fig:network}, the proposed method employs an ensemble-based pipeline with a shared encoder and multiple decoders specialized for different label qualities. The encoder captures shared features across domains and is robust to noise \citep{liu2025}, while branch-specific decoders adapt to high-, mid-, and low-quality supervision.

Since label qualities tend to be geographically distributed---high-quality labels being more prevalent in developed regions such as North America, Europe, and Oceania, while lower-quality labels predominate in less developed areas such as Asia and South America---the label quality inherently encodes domain information. To leverage this, we introduce a "domain" classifier (more precisely, a label quality classifier) positioned atop the shared encoder. This classifier serves as an image classification module, designed to predict the label quality class of the input image.

The encoder and decoder components are flexible and may be instantiated using any suitable monocular height estimation architecture. The domain classifier consists of several convolutional and pooling layers, followed by fully connected layers that produce logits for each domain class, which are then converted into class probabilities $\hat{p}_{c}$.

During inference, the class probabilities are used to weight and blend the height predictions from all branches, yielding the final predicted height map:
\begin{equation}
    \hat{h}=\sum_{c=0}^{C-1}\hat{p}_c\hat{h}^c,
\end{equation}
where $\hat{p}_c\in \mathbb{R}^{C}, \quad \sum_{c=0}^{C-1} \hat{p}_c = 1$ denotes the probability  assigned to domain class $c$ among $C$ total classes, and $\hat{h}^c$ represents the height prediction from branch $c$. During training, however, only the predicted height map from the branch corresponding to the ground truth label quality is used to compute the height-related loss terms.

\subsection{Loss Functions}
To optimally leverage imperfect labels, we design specialized loss functions tailored to each label quality. These loss functions are detailed below.
\subsubsection{"Domain" Classification Loss}
The "domain" classifier is supervised using the "domain" classification loss, a cross-entropy loss, written as
\begin{equation}
    L_{DC}=-\sum_{c=0}^{C-1}y_c \log (\hat{p}_c),
\end{equation}
where $y_c$ is the one-hot ground truth domain label for domain class $c$, and $\hat{p}_c$ is the predicted probability for class $c$.
\subsubsection{Hard Height Loss}
For high-quality labels, which are assumed to be precise, we apply a hard height loss to fully exploit their valuable information. The loss is the standard pixel-wise L1 loss, written as
\begin{equation}
    L_{HH}=\frac{1}{|\mathcal{I}|}\sum_{i \in \mathcal{I}} |\hat{h}_i-h_i|,
\end{equation}
where $\mathcal{I}$ denotes the set of pixels in the image, and $\hat{h}_i$ and $h_i$ are the predicted and ground truth heights at pixel $i$, respectively.
\subsubsection{Balanced Soft Height Loss}
For mid- and low-quality labels, which contain imperfect height information, the supervision signal should be carefully exploited. Specifically, the loss should not penalize predictions that deviate within a certain acceptable margin from the ground truth. To this end, we propose the balanced soft height loss.

First, a buffer zone is defined around the ground truth, within which errors are not penalized. Formally, the per-pixel soft height loss is:
\begin{equation}
    \label{eqn:soft_loss}
    L_{SH} =
        \begin{cases}
            0, & |h_i - \hat{h}_i| \leq \tau \\
            |h_i-\hat{h}_i|, & |h_i - \hat{h}_i| > \tau
        \end{cases},
\end{equation}
where $\tau$ is a pixel-dependent threshold defining the buffer width. In practice, the threshold is set relative to the ground truth height values, that is,
\begin{equation}
    \tau_i=\lambda_{\tau}h_i,
\end{equation}
with $\lambda_{\tau}$ as a hyperparameter controlling the relative threshold.

\begin{figure}[!t]
    \centering
    \includegraphics[width=0.5\linewidth]{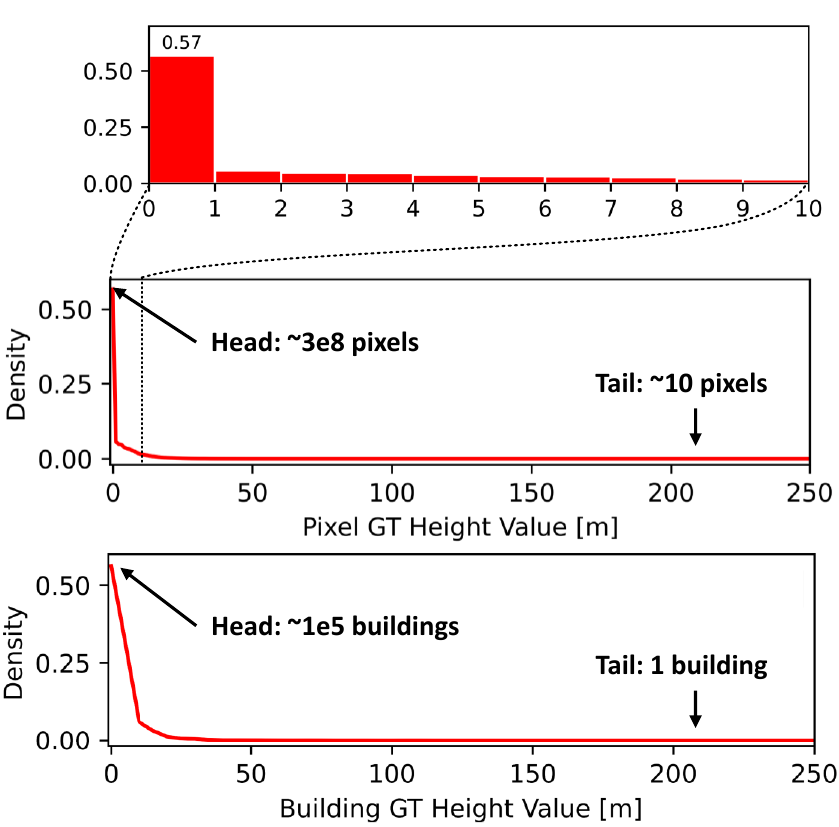}
    \caption{Height value distribution of the GBH high-quality training and validation set. The distribution exhibits a pronounced long-tailed characteristic: approximately 3e8 pixels (57\% of the total) correspond to background regions with heights below 1 m, whereas the number of pixels at large height values decreases sharply to only about 10 per 1 m bin. A similar long-tailed distribution is observed in the building ground-truth height values.}
    \label{fig:long-tail}
\end{figure}

Second, to further soften the supervision and counteract the long-tailed height distribution, the loss is applied only to a sampled subset of pixels $\mathcal{S}\subset\mathcal{I}$. The balanced soft height loss is thus written as
\begin{equation}
    L_{BSH}=\frac{1}{|\mathcal{S}|}\sum_{i \in \mathcal{S}} L_{SH}.
\end{equation}
To ensure balanced sampling that mitigates the effect of imbalanced height values (see Fig. \ref{fig:long-tail}) \citep{chen2023e,chen2023a}, pixels are first sorted by their ground truth height values. A sampling interval is computed by dividing the total number of pixels by the desired sample size, and pixels are then sampled accordingly at regular intervals to form the subset $\mathcal{S}$.
\subsubsection{Ordinal Constraints}
While imperfect labels contain noise in absolute height values, the relative relationships between pixels---specifically, whether one pixel is higher or lower than another---tend to be more reliable. To exploit this, we incorporate an ordinal relation loss as a constraint.

Ordinal relations have been successfully applied in monocular depth estimation \citep{chen2017,lienen2021} and label-efficient knowledge transfer for monocular height estimation \citep{xiong2022}. However, these prior approaches cannot be directly adopted here; careful redesign is required to address the unique challenges of monocular height estimation.

First, to improve the robustness of the ordinal ground truth, we discretize height values into classes using a space-increasing discretization (SID) strategy \citep{fu2018a}. Specifically, the continuous height range $[h_\text{min}, h_\text{max}]$ is partitioned into $K$ classes with boundaries defined as
\begin{equation}
    t_{k}=\exp \!\left(\displaystyle\log (h_\text{min})+\displaystyle\frac{\log(h_\text{max}/h_\text{min})}{K}k\right),
\end{equation}
where $t_k$ is the edge of class $k$. This discretization groups close height values into the same class, reducing the negative impact of ground truth inaccuracies on the ordinal relations.

Second, unlike previous works \citep{chen2017,xiong2022} that also incorporate equality relations, we enforce only strict "higher" or "lower" ordinal relations due to uncertainty in the ground truth. The ordinal constraint loss between pixel 1 and 2 is defined as
\begin{equation}
\label{eqn:ordinal_constraints}
L_{OC}(\hat{h}_1, \hat{h}_2) = 
\begin{cases}
    \log(1 + \exp(\hat{h}_2 - \hat{h}_1)), &c_1 > c_2, \\
    0, & c_1 = c_2, \\
    \log(1 + \exp(\hat{h}_1 - \hat{h}_2)), & c_1 < c_2
\end{cases}
\end{equation}
where $\hat{h}_1$ and $\hat{h}_2$ are predicted heights for the two pixels, and $c_1$ and $c_2$ are their respective ground truth classes.

Third, considering the long-tailed distribution of heights (see Fig. \ref{fig:long-tail}), random sampling of pixel pairs would bias the ordinal constraints toward low-height pixels. To counter this, we adopt a balanced random sampling strategy. Given $M$ height classes, there are 
\begin{equation}
    N_\text{pairs}=\binom{M}{2}=\displaystyle\frac{M(M-1)}{2}
\end{equation}
types of pixel pairs. When sampling a total of $S$ pixel pairs, ideally, each type should have
\begin{equation}
    N_c=\frac{S}{N_\text{pairs}}=\frac{2S}{M(M-1)}
\end{equation}
samples. However, the actual number is limited by the smallest possible pair count across all class pairs. That is,
\begin{equation}
    N_\text{min}=n_{c1}n_{c2},
\end{equation}
where $n_{c1}$ and $n_{c2}$ are the counts of pixels in the two classes with the fewest samples. Thus, for each pair type, we sample 
\begin{equation}
    \hat{N}=\min(N_c,N_\text{min}),
\end{equation}
pairs. This approach ensures a balanced representation of both head and tail classes in ordinal constraint learning.

Finally, since height estimation predicts physical object properties, predicted values should be consistent across different contexts, unlike monocular depth estimation. To enforce this, we include cross-image ordinal relations by flattening and concatenating predicted height maps across images to compute ordinal constraints.
\subsubsection{Total Training Loss}
The total loss $\mathcal{L}$ used to train the pipeline combines all described components: the "domain" classification loss, the hard height loss, the balanced soft height loss, and the ordinal constraints, expressed as
\begin{equation}
    \mathcal{L}=L_{DC}+L_{HH}+L_{BSH}+L_{OC}.
\end{equation}
The loss terms are of the same order of magnitude, so additional scaling is not required to obtain reasonable results.
\subsection{Ground Truth Augmentation}
As observed in experiments described in Section \ref{sec:gt_aug}, incomplete labels lead to over-smoothed predictions from the mid- and low-quality branches. To mitigate this effect, we propose a ground truth augmentation module that generates pseudo-high-quality labels to refine these outputs.

To simulate realistic height maps, fine details must be injected while preserving the accuracy of the original ground truth labels. Predictions from the high-quality branch provide suitable candidates for this purpose, as they often contain more detailed structures and information. Our approach combines these predictions with the original ground truth in a weighted manner to supervise the mid- and low-quality branches.

To reduce the negative impact of erroneous supervisory signals from inaccurate predictions, two safeguards are applied:
\begin{itemize}
    \item \textbf{Error thresholding:} Predictions from the high-quality branch are retained only if their absolute errors are below a threshold.
    \item \textbf{Dropout:} Both original and augmented ground truth are stochastically presented during training to encourage robustness.
\end{itemize}
The augmentation process is defined as
\begin{equation}
    h_\text{aug}=\mathcal{M}\odot \mathcal{H}+(1-\mathcal{M})h,
\end{equation}
where
\begin{equation}
    \mathcal{H}=\eta h+(1-\eta) \hat{h}^0,
\end{equation}
\begin{equation}
\label{eqn:dropout}
\mathcal{M}_i = 
\begin{cases}
    1, & |h_i-\hat{h}_i^0| < \omega \, \text{and not dropout}, \\
    0, & \text{otherwise.} \\
\end{cases}
\end{equation}
Here, $h_\text{aug}$ is the augmented ground truth height map, serving as pseudo-high-quality labels. $h$ is the original ground truth, $\hat{h}^0$ is the prediction from the high-quality branch. The parameter $\eta \in [0,1]$ controls the relative weight of the two height maps and decays over training according to
\begin{equation}
    \eta_{t+1} = \max(\eta_{t}\alpha, 0.5),
\end{equation}
where $\alpha$ is the decay factor and $t$ is the training epoch. The threshold $\omega$ is defined relative to the ground truth value, similar to the definition in the soft height loss.
\section{Experiments}\label{chap:experiments}
\begin{figure*}[!t]
    \centering
    \includegraphics[width=\linewidth]{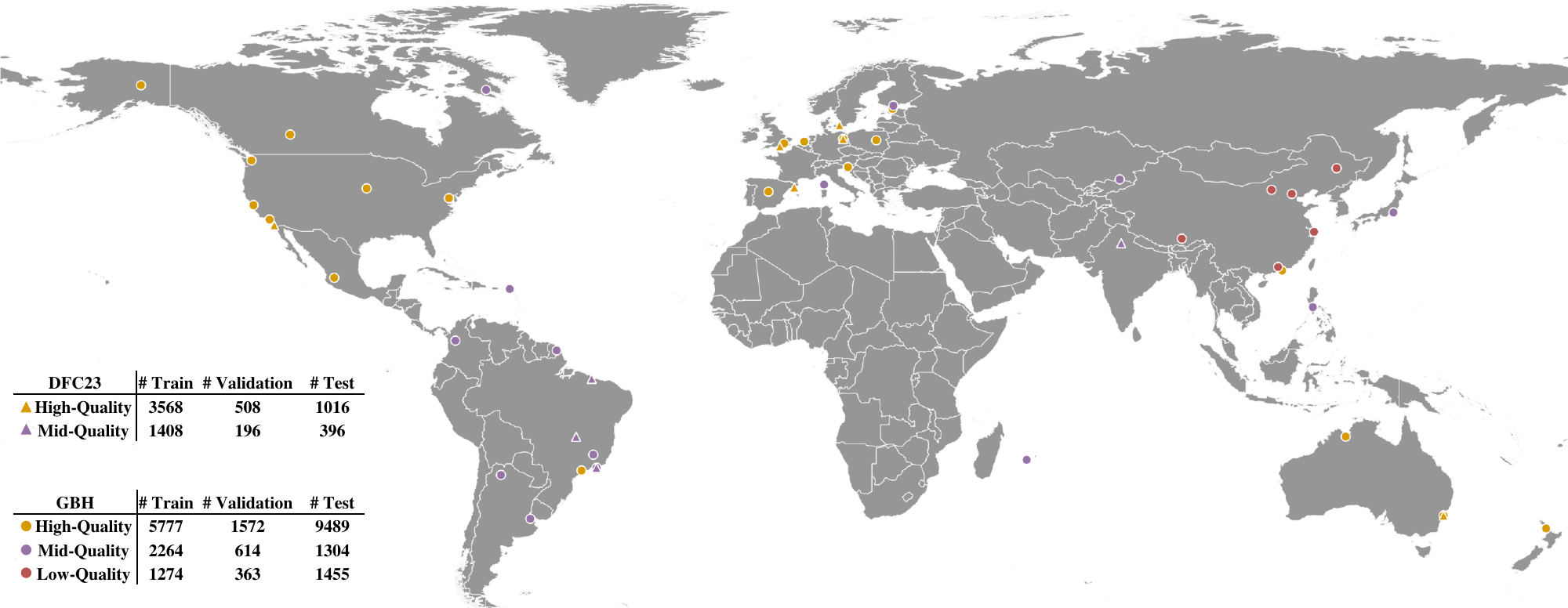}
    \caption{Distribution of data samples.}
    \label{fig:data}
\end{figure*}
\subsection{Datasets}
Two datasets are used to validate the proposed pipeline: DFC23 and GBH, selected for their complementary characteristics. Both contain multi-resolution optical imagery and height annotations of varying quality.
\subsubsection{DFC23}
DFC23 (Data Fusion Contest 2023) dataset \citep{huang2022a,dfc23} was originally designed for large-scale fine-grained building classification in support of semantic urban reconstruction. It contains optical images acquired from SuperView-1 ("GaoJing" in Chinese), Gaofen-2, and Gaofen-3 satellites, along with nDSMs derived from stereo images from Gaofen-7 and WorldView 1/2. The spatial resolutions are 0.5--1 m for the optical images and 2 m for the nDSMs. The dataset is globally distributed, covering 17 cities across six continents. 

It consists of 1,773 image patches of size 512 $\times$ 512 pixels, split into training, validation, and test sets in a 64:16:20 ratio. The patches are further cropped into 7,092 smaller patches of 256 $\times$ 256. The spatial distribution of samples is illustrated in Fig. \ref{fig:data}. 

For this work, the dataset is adapted as follows:
\begin{itemize}
    \item A split is performed by continent: 5,092 patches from North America, Europe, and Oceania are considered to have high-quality labels, while 2,000 patches from other continents are considered to have mid-quality labels.
    \item Mid-quality labels are modified to have a uniform height value for each building instance, with non-building pixels set to zero.
    \item The test set is constructed to include samples from all cities, enabling evaluation across different domains. Specifically, the high-quality test set is treated as the in-domain test set, while data from Brasilia, New Delhi, Sao Luis, and Rio de Janeiro serve as out-of-domain test sets.
\end{itemize}
\subsubsection{GBH}
The GBH (Global Building Height) dataset is a custom-designed dataset that strikes a balance between diversity and volume for monocular height estimation. It consists of optical remote sensing imagery acquired by Planet satellites, height maps generated from open LiDAR observations or building models, and building footprint maps.

% Update the number of patches.
The dataset version used in this work covers 40 cities worldwide from 2013 to 2021. It has a ground sampling distance of 3 m and contains 24,112 patches of size 256 $\times$ 256. Among them, 16,838 patches have high-quality labels, 4,182 patches have mid-quality labels, and 3,092 are annotated with low-quality building height proxies, such as building floor counts. An overview is shown in Fig. \ref{fig:data}.

The dataset is used as follows:
\begin{itemize}
    \item The dataset is split into high-, mid-, and low-quality sets. Each set is further divided into training, validation, and test subsets.
    \item Labels are already provided in their respective quality levels, so no further label processing is required.
    \item Four cities---Los Angeles (USA), Sao Paulo (Brazil), Medellin (Colombia), and Guangzhou (China)---are held out entirely for testing and never seen during training.
    \item Test sets are categorized as in-domain or out-of-domain based on geographic location and quality. Specifically, the high-quality test set and Los Angeles constitute the in-domain test sets, while the mid-/low-quality test set, Sao Paulo, Medellin, and Guangzhou form the out-of-domain test sets.
\end{itemize}

\subsection{Evaluation Metrics}
The models are evaluated using the building-wise root mean square error (RMSE). Due to the low quality of certain labels---specifically, the presence of only valid but oversmoothed heights for building pixels---the evaluation is valid only under the assumption of LoD1 building models for each building instance. In other words, only the median height value within each building footprint is considered. For each experiment, the average building-wise RMSE is computed for the in-domain, out-of-domain, and combined test sets, enabling the assessment of the model's generalization capability across diverse domains.

\subsection{Implementation Details}
We validate the proposed pipeline using four baselines: U-Net \citep{ronneberger2015}, IM2HEIGHT \citep{mou2018}, AdaBins \citep{bhat2021}, and DepthAnything v2 \citep{depthanything}. U-Net serves as a generic fully convolutional model; IM2HEIGHT is an early task-specific network; AdaBins underpins the recent HTC-DC Net \citep{chen2023e}; and DepthAnything v2 represents a transformer-based depth foundation model. The U-Net is adapted by removing its final activation layer.

Each baseline is trained (1) on high-quality data only, (2) on mid- or mid/low-quality data only, (3) on all data jointly, and (4) within our full pipeline using the baseline as backbone.

For DFC23, the ensemble comprises high- and mid-quality branches; for GBH, an additional branch handles low-quality labels (converted to meters assuming 3 m per floor).

Soft height loss thresholds are set to 0.3× and 0.5× the ground truth height for mid- and low-quality labels, respectively, computed on 10\% sampled pixels. Ordinal constraints utilize 20 discrete classes over $H\times W$ pixel pairs per image. Ground truth augmentation accepts predictions within 0.1× ground truth error, applies dropout = 0.3, and decays $\eta$ by 0.99 per epoch.

All models are trained for 200 epochs using the Adam optimizer (learning rate = 1e-4), and the best validation epoch is reported across three independent runs.
\section{Results}\label{chap:results}
Quantitative results are reported in Tables \ref{tab:res_dfc} and \ref{tab:res_gbh}, while qualitative comparisons are shown in Figs. \ref{fig:res_vis_dfc} and \ref{fig:res_vis_gbh}. Overall, the proposed pipeline consistently improves performance in regions with imperfect labels, while maintaining comparable results in areas with high-quality labels.
\subsection{DFC23}
\begin{table*}
    \centering
    \caption{Experimental results on DFC23. Building-wise RMSEs are reported in meters. Each experiment is conducted three times, and the average metrics are presented. H: High-Quality; M: Mid-Quality. The best result in each set is highlighted in \firstbest{green}, and the second-best in \secondbest{blue}.}
    \resizebox{\linewidth}{!}{
    \begin{tabular}{cc|c|cccc|c|c}
    \toprule
         \multirow{2}{*}{Network} & \multirow{2}{*}{Trained on} & In-domain & \multicolumn{5}{c|}{Out-of-domain} & \multirow{2}{*}{Average} \\
          &  & High-Quality Test Set & Brasilia & New Delhi & Sao Luis & Rio de Janeiro & Average & \\
         \midrule
            \multirow{3}{*}{U-Net \citep{ronneberger2015}} & H & \firstbest{4.3308} & 6.2238 & 11.1197 & 6.8789 & 13.4260 & 9.4121 & 8.3958 \\ 
            ~ & M & 15.0725 & \firstbest{1.8674} & \firstbest{4.3610} & \firstbest{2.0855} & \firstbest{7.2211} & \firstbest{3.8837} & 6.1215 \\ 
            ~ & H+M & 5.3343 & 2.8716 & 4.8708 & 2.6416 & 8.9380 & 4.8305 & \secondbest{4.9312} \\ \midrule
            ours + U-Net & H+M & \secondbest{4.8461} & \secondbest{2.4613} & \secondbest{4.7726} & \secondbest{2.6402} & \secondbest{7.9878} & \secondbest{4.4655} & \firstbest{4.5416} \\ \midrule \midrule
            \multirow{3}{*}{IM2HEIGHT \citep{mou2018}} & H & \firstbest{5.8326} & 5.1242 & 10.6451 & 7.3185 & 13.9272 & 9.2538 & 8.5695 \\ 
            ~ & M & 14.8098 & 2.4793 & 5.6552 & 2.5868 & \secondbest{8.0383} & \secondbest{4.6899} & 6.7139 \\ 
            ~ & H+M & \secondbest{5.9918} & \secondbest{2.4590} & \secondbest{5.6141} & \secondbest{2.5631} & 8.4025 & 4.7597 & \secondbest{5.0061} \\ \midrule 
            ours + IM2HEIGHT & H+M & 6.0651 & \firstbest{1.8587} & \firstbest{3.9865} & \firstbest{2.1776} & \firstbest{7.3389} & \firstbest{3.8404} & \firstbest{4.2853} \\  \midrule \midrule
            \multirow{3}{*}{AdaBins B5 \citep{bhat2021}} & H & \firstbest{4.2442} & 6.5472 & 9.9179 & 7.1489 & 12.4282 & 9.0105 & 8.0573 \\ 
            ~ & M & 13.2865 & \secondbest{2.2143} & \secondbest{4.8853} & 2.5159 & \secondbest{7.5285} & \secondbest{4.2860} & 6.0861 \\ 
            ~ & H+M & 5.9559 & 2.6537 & 5.6449 & \secondbest{2.4535} & 10.3187 & 5.2677 & \secondbest{5.4053} \\ \midrule
            ours + AdaBins B5 & H+M & \secondbest{5.6022} & \firstbest{1.7491} & \firstbest{3.9914} & \firstbest{2.1789} & \firstbest{7.3040} & \firstbest{3.8059} & \firstbest{4.1652} \\ 
            \midrule \midrule
            \multirow{3}{*}{DepthAnything v2 S \citep{depthanything}} & H & \secondbest{6.6766} & 8.8032 & 8.6486 & 11.4328 & 12.5349 & 10.3549 & 9.6192 \\ 
            ~ & M & 13.7458 & 3.1552 & 7.1507 & \secondbest{3.4726} & 12.3118 & 6.5226 & 7.9672 \\ 
            ~ & H+M & 6.8089 & \secondbest{2.9916} & \secondbest{6.6290} & 3.9013 & \secondbest{11.6931} & \secondbest{6.3037} & \secondbest{6.4048} \\ \midrule
            ours + DepthAnything v2 S & H+M & \firstbest{5.9478} & \firstbest{2.5207} & \firstbest{5.1246} & \firstbest{2.9778} & \firstbest{9.2009} & \firstbest{4.9560} & \firstbest{5.1543} \\ 
        \bottomrule
    \end{tabular}}
    \label{tab:res_dfc}
\end{table*}
\begin{figure*}[!ht]
    \centering
    \includegraphics[width=0.7\linewidth]{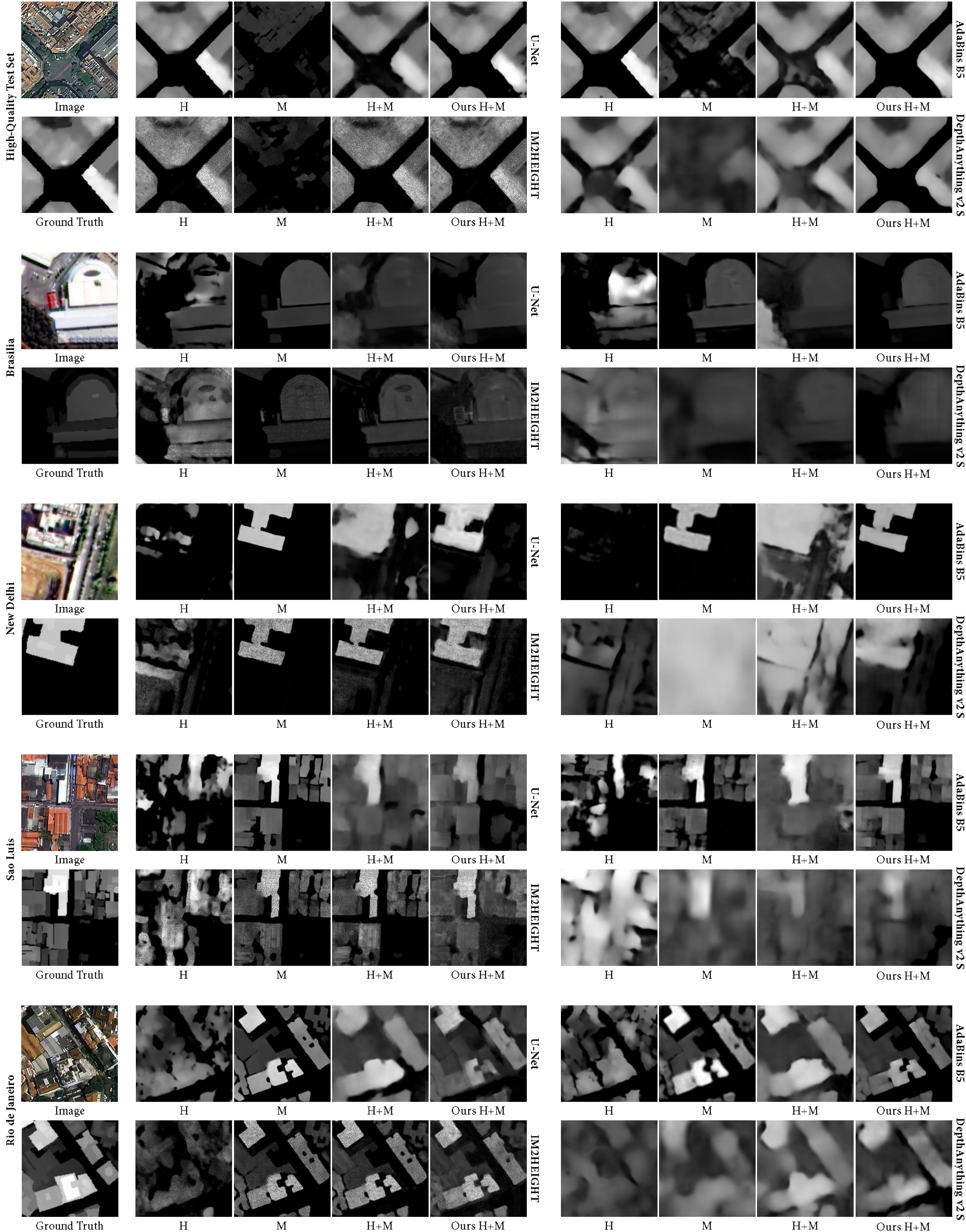}
    \caption{Qualitative results of different models on DFC23. The maps are scaled to the same range. One sample from each test set is visualized per block. H: Backbone network trained on high-quality labels only; M: Backbone network trained on mid-quality labels only; H+M: Backbone network trained on both high-quality and mid-quality labels; Ours H+M: Our proposed pipeline using the backbone network trained on both high- and mid-quality labels.}
    \label{fig:res_vis_dfc}
\end{figure*}
As shown in Table \ref{tab:res_dfc}, our proposed pipeline consistently enhances the performance of the backbone networks, achieving the lowest average building-wise RMSE across all configurations. Specifically, relative to the best baseline (H+M), average RMSE is reduced by 7.90\%, 14.40\%, 22.94\%, and 19.52\% for U-Net, IM2HEIGHT, AdaBins B5, and DepthAnything v2 S, respectively. Performance remains comparable in high-quality regions but improves significantly in cities with mid-quality labels, indicating that imperfect supervision effectively boosts generalization.

These quantitative gains are further supported by the qualitative results in Fig. \ref{fig:res_vis_dfc}. Our approach yields sharper building boundaries and clearer structural patterns, particularly in mid-quality regions where conventional training often produces blurred results.

It is noteworthy that DepthAnything v2 S, despite being a state-of-the-art depth foundation model, performs worse than a simple U-Net backbone in our experiments. Transformer-based architectures typically require large-scale datasets to reach their full capacity, emphasizing global context over local detail, which can lead to blurred building boundaries. They also depend on dense, high-quality supervision, making them less robust to noisy labels. Nevertheless, our pipeline alleviates these limitations by sharpening boundaries and compensating for the weaker local inductive bias of transformers.
\subsection{GBH}
\begin{table*}[t]
    \centering
    \caption{Experimental results on GBH. L: Low-Quality. See Table \ref{tab:res_dfc} for other notations.}
    \resizebox{\linewidth}{!}{
    \begin{tabular}{cc|cc|c|cccc|c|c}
    \toprule
        \multirow{3}{*}{Network} & \multirow{3}{*}{Trained on} & \multicolumn{3}{c|}{In-domain} & \multicolumn{5}{c|}{Out-of-domain} & \multirow{3}{*}{Average} \\
          &  & High-Quality & Los & \multirow{2}{*}{Average} & Mid-/Low-Quality & Sao & \multirow{2}{*}{Medellin} & \multirow{2}{*}{Guangzhou} & \multirow{2}{*}{Average} & \\
         & & Test Set & Angeles & & Test Set & Paulo & & & \\
         \midrule
            \multirow{3}{*}{U-Net \citep{ronneberger2015}} & H & \firstbest{3.8915} & \firstbest{2.4214} & \firstbest{3.1564} & 16.0513 & \secondbest{6.3528} & \secondbest{4.3093} & \secondbest{13.7006} & 10.1035 & 7.7878 \\ 
            ~ & M/L & 7.6104 & 7.8630 & 7.7367 & \secondbest{11.6371} & 6.4807 & 4.6644 & 15.7384 & 9.6301 & 8.9990 \\ 
            ~ & H+M/L & \secondbest{4.0981} & 3.6468 & \secondbest{3.8724} & 11.7305 & 6.3704 & 4.7381 & 14.5750 & \secondbest{9.3535} & \secondbest{7.5265} \\ \midrule
            ours + U-Net & H+M/L & 4.2323 & \secondbest{3.5862} & 3.9093 & \firstbest{9.2725} & \firstbest{5.5070} & \firstbest{3.8394} & \firstbest{12.5761} & \firstbest{7.7987} & \firstbest{6.5023} \\ \midrule \midrule
            \multirow{3}{*}{IM2HEIGHT \citep{mou2018}} & H & \firstbest{5.1190} & \firstbest{2.9118} & \firstbest{4.0154} & 17.6651 & \secondbest{6.8923} & \secondbest{4.4851} & \secondbest{16.8945} & 11.4843 & 8.9946 \\ 
            ~ & M/L & 8.1578 & 4.5641 & 6.3610 & 15.9578 & 7.7574 & 5.3270 & 17.9398 & 11.7455 & 9.9506 \\ 
            ~ & H+M/L & \secondbest{5.1893} & 3.2747 & \secondbest{4.2320} & \secondbest{14.8351} & 7.4922 & 5.2060 & 17.3906 & \secondbest{11.2310} & \secondbest{8.8980} \\ \midrule
            ours + IM2HEIGHT & H+M/L & 5.6562 & \secondbest{3.0636} & 4.3599 & \firstbest{10.5633} & \firstbest{6.0087} & \firstbest{3.9307} & \firstbest{14.2240} & \firstbest{8.6817} & \firstbest{7.2411} \\ \midrule \midrule
            \multirow{3}{*}{AdaBins B5 \citep{bhat2021}} & H & \secondbest{4.0148} & \firstbest{2.2375} & \firstbest{3.1261} & 16.2250 & \secondbest{6.4674} & \secondbest{4.3789} & \secondbest{13.8260} & 10.2243 & 7.8583 \\ 
            ~ & M/L & 6.9874 & 4.6826 & 5.8350 & 11.8719 & 6.5632 & 4.6940 & 15.4551 & 9.6460 & 8.3757 \\ 
            ~ & H+M/L & \firstbest{3.9835} & 2.6641 & \secondbest{3.3238} & \secondbest{11.7416} & 6.5350 & 4.7810 & 14.3911 & \secondbest{9.3622} & \secondbest{7.3494} \\ \midrule
            ours + AdaBins B5 & H+M/L & 4.5589 & \secondbest{2.5139} & 3.5364 & \firstbest{9.2164} & \firstbest{5.2202} & \firstbest{3.9779} & \firstbest{12.2288} & \firstbest{7.6608} & \firstbest{6.2860} \\ 
            \midrule \midrule
        \multirow{3}{*}{DepthAnything v2 S \citep{depthanything}} & H & \firstbest{4.6917} & \firstbest{3.0083} & \firstbest{3.8500} & 17.2878 & \secondbest{7.0030} & \secondbest{4.7398} & \secondbest{16.2738} & \secondbest{9.3389} & 8.8341 \\
        ~ & M/L & 8.2570 & 6.2220 & 7.2395 & 15.4904 & 7.4221 & 5.2248 & 16.9222 & 9.8564 & 9.9231 \\
        ~ & H+M/L & \secondbest{4.9024} & 3.5098 & \secondbest{4.2061} & \secondbest{14.6202} & 7.1862 & 4.9761 & 16.3485 & 9.5036 & \secondbest{8.5905} \\ \midrule
        ours + DepthAnything v2 S & H+M/L & 5.3508 & \secondbest{3.4369} & 4.3938 & \firstbest{11.1049} & \firstbest{6.1256} & \firstbest{4.3104} & \firstbest{12.8607} & \firstbest{7.7655} & \firstbest{7.1982} \\ 
        \bottomrule
    \end{tabular}}
    \label{tab:res_gbh}
\end{table*}
\begin{figure*}[!ht]
    \centering
    \includegraphics[width=0.7\linewidth]{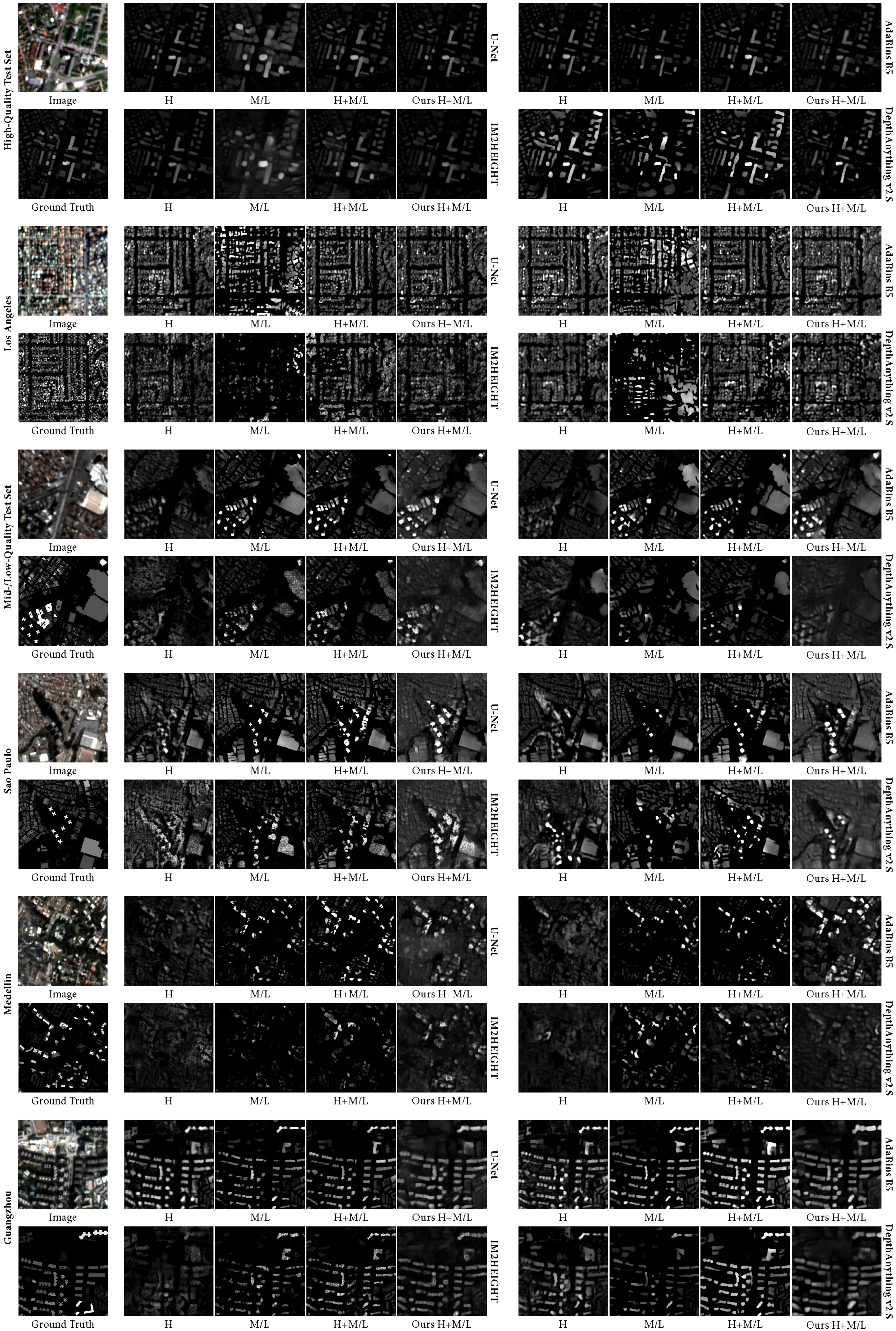}
    \caption{Qualitative results of different models on GBH. L: Low-quality. See Fig. \ref{fig:res_vis_dfc} for other notations.}
    \label{fig:res_vis_gbh}
\end{figure*}
The GBH dataset covers a broader range of cities, including those with low-quality labels, enabling evaluation under challenging conditions. As reported in Table \ref{tab:res_gbh}, our pipeline significantly outperforms all baselines (H+M/L), reducing average RMSE by 13.61\%, 18.62\%, 14.47\%, and 16.21\% for U-Net, IM2HEIGHT, AdaBins B5, and DepthAnything v2 S, respectively. The largest gains occur in cities with only mid- or low-quality labels, compensating for minor drops in high-quality areas and in Los Angeles.

The qualitative results in Fig. \ref{fig:res_vis_gbh} further support these findings, showing that our pipeline preserves fine-grained structural details and generates clearer building boundaries. These visualizations also indicate that our approach effectively integrates the complementary strengths of networks trained on different subsets of the training data.

In summary, the proposed pipeline leverages imperfect labels to achieve significantly better performance in cities with lower-quality annotations while maintaining comparable results elsewhere. This results in a more balanced overall performance across various domains. Together, the results highlight the robustness and adaptability of our method across heterogeneous urban environments and varying label qualities.
\section{Discussion} \label{chap:discussions}
\subsection{Model Generalizability with Different Training Data}
Our experiments with four monocular height estimation networks demonstrate that model generalization depends strongly on both the quality and diversity of the training data. Models trained solely on high-quality labels perform well within their domain but generalize poorly to other domains, often producing noisy or spurious patterns on out-of-domain images. Conversely, models trained only on mid- or low-quality labels suffer from label noise, yielding inaccurate height predictions across all regions.

Training with mixed-quality data enhances diversity but can still compromise performance when noisy supervision prevails. The proposed multi-branch architecture and tailored loss functions mitigate this effect by selectively extracting reliable signals from imperfect labels, resulting in more balanced cross-domain performance compared to naive mixed training.

Overall, our findings emphasize that data quality and diversity remain the primary determinants of robust generalization in monocular height estimation and that properly leveraging imperfect supervision can substantially extend model applicability across diverse geographic domains.
\subsection{Ablation Studies}
We conduct extensive ablation experiments to evaluate the contribution of each design component in our proposed pipeline. All experiments are conducted on the DFC23 and GBH datasets, utilizing the U-Net as the backbone.

Table \ref{tab:abl_dfc} and \ref{tab:abl_gbh} report building-wise RMSEs for various configurations. The evaluated components include:
\begin{itemize}
    \item Ens.: multi-branch ensemble.
    \item $L_{DC}$: "domain" classifier.
    \item $L_{BSH}$: balanced soft height loss.
    \item $L_{OC}$: ordinal constraints.
    \item GT Aug.: ground truth augmentation.
\end{itemize}

\begin{table*}[!ht]
    \centering
    \caption{Ablation studies on DFC23. Building-wise RMSEs are reported in meters. The pipeline is backboned by U-Net. Ens.: Ensemble; $L_{DC}$: Domain Classifier; $L_{BSH}$: Balanced Soft Height Loss; $L_{OC}$: Ordinal Constraints; GT Aug.: Ground Truth Augmentation. The best is highlighted in \firstbest{green}.}
    \resizebox{\linewidth}{!}{
    \begin{tabular}{ccccc|c|cccc|c|c}
    \toprule
         \multirow{2}{*}{Ens.} & \multirow{2}{*}{$L_{DC}$} & \multirow{2}{*}{$L_{BSH}$} & \multirow{2}{*}{$L_{OC}$} & GT & In-domain & \multicolumn{5}{c|}{Out-of-domain} & \multirow{2}{*}{Average} \\
         & & & & Aug. & High-Quality Test Set & Brasilia & New Delhi & Sao Luis & Rio de Janeiro & Average & \\
         \midrule
         \checkmark &  &  &  &  & 7.4842 & 9.3961 & 8.7850 & 6.2896 & 11.4998 & 8.9926 & 8.6909 \\ 
         \checkmark & \checkmark & & & & 4.9835 & 2.3857 & 4.8198 & \firstbest{2.2624} & 9.3292 & 4.6993 & 4.7562 \\
         \checkmark & \checkmark & \checkmark & & & 5.0231 & 3.1498 & 5.7679 & 2.6543 & 8.6232 & 5.0488 & 5.0437 \\
         \checkmark & \checkmark & \checkmark & \checkmark & & \firstbest{4.6485} & \firstbest{2.1197} & 4.8313 & 2.3912 & 8.8752 & 4.5544 & 4.5732 \\ 
         \checkmark & \checkmark & & & \checkmark & 5.1768 & 4.0525 & \firstbest{4.5670} & 2.3442 & \firstbest{7.6107} & 4.6436 & 4.7502 \\ \midrule
         \checkmark & \checkmark & \checkmark & \checkmark & \checkmark & 4.8461 & 2.4613 & 4.7726 & 2.6402 & 7.9878 & \firstbest{4.4655} & \firstbest{4.5416} \\ 
    \bottomrule
    \end{tabular}
    \label{tab:abl_dfc}}
\end{table*}
\begin{table*}[!ht]
    \centering
    \caption{Ablation studies on GBH. See Table \ref{tab:abl_dfc} for notations.}
    \resizebox{\linewidth}{!}{
    \begin{tabular}{ccccc|cc|c|cccc|c|c}
    \toprule
         \multirow{3}{*}{Ens.} & \multirow{3}{*}{$L_{DC}$} & \multirow{3}{*}{$L_{BSH}$} & \multirow{3}{*}{$L_{OC}$} & \multirow{2}{*}{GT} & \multicolumn{3}{c|}{In-domain} & \multicolumn{5}{c|}{Out-of-domain} & \multirow{3}{*}{Average} \\
          & & & & & High-Quality & Los & \multirow{2}{*}{Average} & Mid-/Low-Quality & Sao & \multirow{2}{*}{Medellin} & \multirow{2}{*}{Guangzhou} & \multirow{2}{*}{Average} & \\
         & & & & Aug. & Test Set & Angeles & & Test Set & Paulo & & & \\
         \midrule
         \checkmark & ~ & ~ & ~ & & 5.8309 & 5.4246 & 5.6277 & 11.4728 & 6.1393 & 4.6286 & \firstbest{11.6786} & 8.4798 & 7.5291 \\ 
         \checkmark & \checkmark & ~ & ~ & & 3.7807 & 9.7912 & 6.7859 & 8.4418 & 5.5307 & 3.8394 & 12.1407 & \firstbest{7.4881} & 7.2541 \\ 
         \checkmark & \checkmark & \checkmark & & & 4.2620 & 4.5810 & 4.4215 & 9.3561 & \firstbest{5.4464} & 4.0296 & 12.4383 & 7.8176 & 6.6856 \\
         \checkmark & \checkmark & \checkmark & \checkmark & & 4.3612 & 3.8057 & 4.0835 & 8.7765 & 5.4832 & 4.0373 & 13.0587 & 7.8389 & 6.5871 \\ 
         \checkmark & \checkmark &  & ~ & \checkmark & \firstbest{3.6792} & 7.0112 & 5.3452 & \firstbest{8.3891} & 5.5962 & 6.2899 & 13.0767 & 8.3380 & 7.3404 \\ \midrule
         \checkmark & \checkmark & \checkmark & \checkmark & \checkmark & 4.2323 & \firstbest{3.5862} & \firstbest{3.9093} & 9.2725 & 5.5070 & \firstbest{3.8394} & 12.5761 & 7.7987 & \firstbest{6.5023} \\ 
    \bottomrule
    \end{tabular}}
    \label{tab:abl_gbh}
\end{table*}
\subsubsection{"Domain" Classifier}
The "domain" classifier is designed to identify underlying domains implicitly reflected in the label quality distribution. It is trained jointly with the height predictors, enabling the incorporation of domain-specific cues during training. At inference time, it also guides the aggregation of multi-branch outputs by assigning higher weights to branches aligned with the predicted domain.

As shown in Table \ref{tab:abl_dfc} and Table \ref{tab:abl_gbh}, the "domain" classifier yields notable performance gains, particularly on out-of-domain test sets. On DFC23, the average building-wise RMSE decreases from 8.6909 m to 4.7562 m, a 45.27\% improvement. On GBH, the average building-wise RMSE improves from 7.5291 m to 7.2541 m, corresponding to a 3.65\% improvement, with the most pronounced gains of 11.69\% on out-of-domain subsets. These results demonstrate that the "domain" classifier effectively leverages implicit domain cues to enhance cross-domain generalization. 
\subsubsection{Balanced Soft Height Loss}
The balanced soft height loss leverages imperfect labels by filtering the supervisory signal based on absolute height values. This design addresses two key issues: (1) noise in imperfect labels that can degrade model performance, and (2) the imbalanced distribution of height values. 

As shown in Table \ref{tab:abl_dfc} and Table \ref{tab:abl_gbh}, the balanced soft height loss alone does not consistently improve overall performance. However, it effectively mitigates the adverse effects of noisy labels, leading to more stable results---particularly in Los Angeles. When combined with the other proposed modules, the performance gains become substantial. These results underscore the importance of explicitly accounting for annotation reliability and label distribution imbalance in the loss design.
\begin{table*}[!ht]
    \centering
    \caption{Ablation study of hyperparameters for the balanced soft height loss on GBH. Building-wise RMSEs are reported in meters. The pipeline is backboned by U-Net. $\alpha$ is the sampling ratio, and $\tau$ is the threshold. The subscripts "1" and "2" denote mid-quality and low-quality labels, respectively. Random sampling means no balanced sampling is conducted.}
    \resizebox{\linewidth}{!}{
    \begin{tabular}{cccc|cc|c|cccc|c|c}
    \toprule
         \multirow{3}{*}{$\alpha_1$} & \multirow{3}{*}{$\tau_1$} & \multirow{3}{*}{$\alpha_2$} & \multirow{3}{*}{$\tau_2$} & \multicolumn{3}{c|}{In-domain} & \multicolumn{5}{c|}{Out-of-domain} & \multirow{3}{*}{Average} \\
          & & & & High-Quality & Los & \multirow{2}{*}{Average} & Mid-/Low-Quality & Sao & \multirow{2}{*}{Medellin} & \multirow{2}{*}{Guangzhou} & \multirow{2}{*}{Average} & \\
         & & & & Test Set & Angeles & & Test Set & Paulo & & &  \\
         \midrule
        0.1 & 0.3 & 0.1 & 0.5 & 4.2323 & 3.5862 & 3.9093 & 9.2725 & 5.5070 & 3.8394 & 12.5761 & 7.7987 & 6.5023 \\ \midrule

        0.3 & 0.3 & 0.1 & 0.5 & 4.4870 & 4.2990 & 4.3930 & 8.5110 & 5.7529 & 3.6607 & 13.1274 & 7.7630 & 6.6397 \\ 
        0.5 & 0.3 & 0.1 & 0.5 & 4.3925 & 3.6758 & 4.0341 & 9.2499 & 5.5590 & 4.5100 & 13.3259 & 8.1612 & 6.7855 \\ \midrule

        0.1 & 0.1 & 0.1 & 0.5 & 4.1332 & 3.3172 & 3.7252 & 8.8506 & 5.4144 & 3.7973 & 13.1086 & 7.7927 & 6.4369 \\ 
        0.1 & 0.5 & 0.1 & 0.5 & 6.3059 & 4.6073 & 5.4566 & 15.2756 & 6.3538 & 4.0265 & 14.5987 & 10.0636 & 8.5280 \\ \midrule
        
        0.1 & 0.3 & 0.3 & 0.5 & 4.1340 & 3.0466 & 3.5903 & 9.0064 & 5.3923 & 3.9559 & 12.6697 & 7.7561 & 6.3675 \\ 
        0.1 & 0.3 & 0.5 & 0.5 & 4.2786 & 3.7872 & 4.0329 & 9.6824 & 5.4229 & 3.9600 & 13.2812 & 8.0866 & 6.7354 \\  \midrule
        
        0.1 & 0.3 & 0.1 & 0.3 & 4.1514 & 3.2231 & 3.5938 & 8.9971 & 5.3609 & 3.9369 & 12.3631 & 7.6645 & 6.3388 \\
        0.1 & 0.3 & 0.1 & 0.7 & 4.2855 & 2.9020 & 3.6873 & 9.8572 & 5.5934 & 3.7270 & 13.6117 & 8.1973 & 6.6628 \\ 

        \midrule
        \multicolumn{4}{c|}{Random Sampling} & 4.2257 & 4.3518 & 4.2887 & 8.9230 & 5.4780 & 3.9432 & 13.7806 & 8.0312 & 6.7837 \\
        \bottomrule
    \end{tabular}}
    \label{tab:bsh_gbh}
\end{table*}

We further examine the impact of the sampling ratio $\alpha$ and threshold $\tau$ in the balanced soft height loss on the GBH dataset (see Table \ref{tab:bsh_gbh}). Overall, performance remains relatively stable across configurations, with the default settings near optimal.

For mid-quality labels, a slight increase in the sampling rate ($\alpha_1$) results in degraded performance. Tightening the boundaries (i.e., decreasing $\tau_1$) marginally improves the results, although this introduces more noise from the noisy labels. Relaxing the boundaries (i.e., increasing $\tau_1$) removes noise but sacrifices functional supervision, resulting in consistently worse performance.

For low-quality labels, moderate increases in sampled pixels ($\alpha_2$) provide marginal benefits before plateauing. Narrowing the buffer zone (i.e., decreasing $\tau_2$) generally improves results, whereas broadening it (i.e., increasing $\tau_2$) harms performance---most notably on out-of-domain test sets.

Replacing the balanced sampling strategy with random sampling noticeably reduces accuracy, confirming the value of explicitly addressing imbalanced height distributions through targeted sampling.
\subsubsection{Ordinal Constraints}
The ordinal constraints encourage the model to produce predictions consistent with relative height relationships, providing a complementary supervisory signal to the balanced soft loss. These constraints are derived from noisy height annotations, where relative ordering is often more reliable than the absolute height values. As shown in Table \ref{tab:abl_dfc} and Table \ref{tab:abl_gbh}, incorporating ordinal constraints significantly improves performance---enhancing in-domain performance and achieving a more balanced trade-off between in-domain and out-of-domain results.
\begin{table*}[!ht]
    \centering
    \caption{Ablation study of ordinal constraints on GBH. Building-wise RMSEs are reported in meters. The pipeline is backboned by U-Net. Constraint Types: (H+M)C+L: cross-image constraints for high- and mid-quality labels, within-image constraints for low-quality labels; H+M: within-image constraints for high- and mid-quality labels; (H+M)C: high- and mid-quality labels, cross-image constraints; H+M+L: within-image constraints for high-, mid-, and low-quality labels. Random sampling: No balanced sampling of pixel pairs is conducted.}
    \resizebox{\linewidth}{!}{
    \begin{tabular}{c|cc|c|cccc|c|c}
    \toprule
     \multirow{2}{*}{Constraints Type} & \multicolumn{3}{c|}{In-domain} & \multicolumn{5}{c|}{Out-of-domain} & \multirow{2}{*}{Average} \\
     & High-Quality Test Set & Los Angeles & Average & Mid-/Low-Quality Test Set & Sao Paulo & Medellin & Guangzhou & Average & \\ \midrule
        (H+M)C+L & 4.2323 & 3.5862 & 3.9093 & 9.2725 & \firstbest{5.5070} & 3.8394 & 12.5761 & \firstbest{7.7987} & \firstbest{6.5023} \\ \midrule
        H+M & 4.4211 & 4.1325 & 4.2768 & 9.3354 & 5.5368 & 4.2871 & \firstbest{12.4416} & 7.9002 & 6.6924 \\ 
        (H+M)C & 4.2416 & 3.6140 & 3.9278 & 9.1870 & 5.7148 & 6.0226 & 13.2404 & 8.5412 & 7.0034 \\ 
        H+M+L & \firstbest{4.1428} & \firstbest{2.9659} & \firstbest{3.5543} & 9.4912 & 5.6216 & 3.8501 & 13.2327 & 8.0489 & 6.5507 \\ \midrule
        Random Sampling & 4.4316 & 6.5894 & 5.5105 & \firstbest{8.7607} & 5.7695 & \firstbest{3.8065} & 12.8886 & 7.8063 & 7.0410 \\ 
        \bottomrule
    \end{tabular}}
    \label{tab:cons_gbh}
\end{table*}

We conduct additional experiments summarized in Table \ref{tab:cons_gbh} to further explore the effect of different strategies for enforcing ordinal constraints. The results indicate that incorporating low-quality labels when computing ordinal constraints is beneficial for enhancing overall robustness, particularly in improving in-domain results. This suggests that even relatively low-quality labels can provide valuable relative information to guide learning.

Additionally, leveraging cross-image ordinal relations can yield further performance gains, particularly in out-of-domain cities, by enabling the model to generalize relative height relationships across different scenes.

All ordinal constraints are computed using a balanced sampling strategy, where pixel pairs are uniformly sampled across height differences. When this strategy is replaced by random sampling, in-domain performance drops substantially. 
\subsubsection{Ground Truth Augmentation}\label{sec:gt_aug}
\begin{figure*}[!ht]
    \centering
    \includegraphics[width=\linewidth]{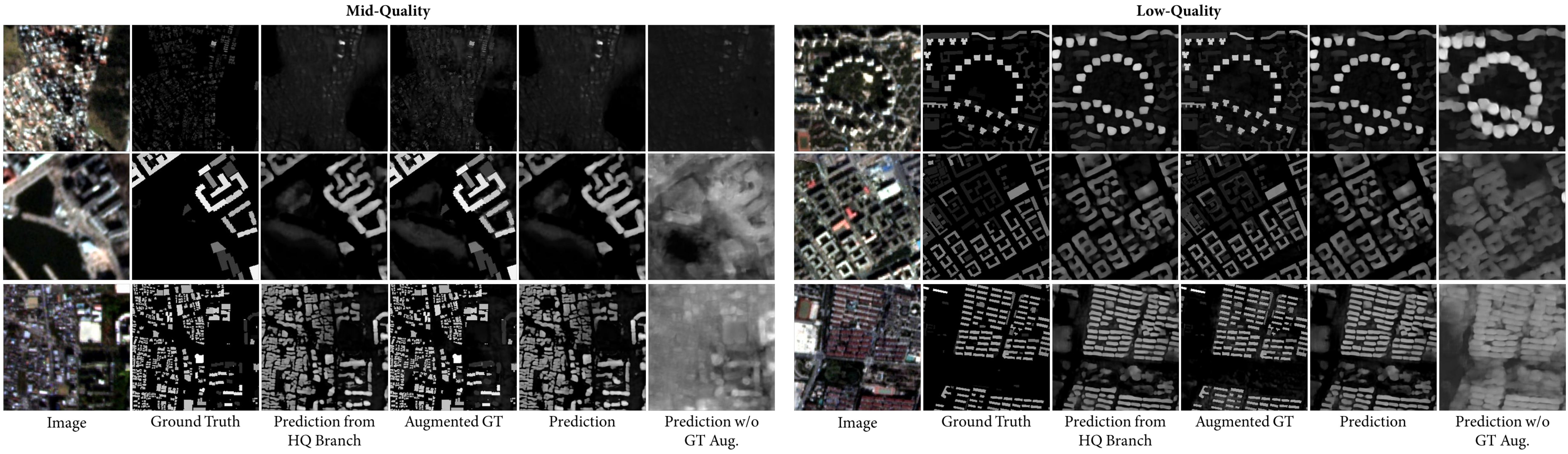}
    \caption{Visualization of the deblurring effect from ground truth augmentation. The maps are scaled to the same range.}
    \label{fig:abl_gt_aug}
\end{figure*}

Quantitative improvements do not always align with perceptual gains. As shown in Tables \ref{tab:abl_dfc} and \ref{tab:abl_gbh}, adding all components, ensemble learning, domain classifier, balanced soft height loss, and ordinal constraints, yields substantially better metrics than baselines, yet the resulting predictions appear overly smoothed (see Fig. \ref{fig:abl_gt_aug}). This occurs because mid- and low-quality labels resemble blurred ground truths, prompting the network to reproduce such patterns.

To counter this, we apply ground truth augmentation, which fuses reliable predictions from the high-quality branch to refine mid- and low-quality annotations, producing sharper and more detailed supervision. The resulting models generate outputs with finer structural detail and improved visual fidelity (see Fig. \ref{fig:abl_gt_aug}).

During training, we introduce dropout on the augmentation path to prevent overfitting and enhance the flexibility of using the augmented ground truth maps. This means the augmentation is probabilistically disabled during training. As illustrated in Table \ref{tab:do_gbh}, varying this probability can alter model performance, with a dropout probability of 0.3 yielding the best results.
\begin{table*}[!ht]
    \centering
    \caption{Ablation study of ground truth augmentation dropout probability on GBH. Building-wise RMSEs are reported in meters. The pipeline is backboned by U-Net. The best is highlighted in \firstbest{green}.}
    \resizebox{\linewidth}{!}{
    \begin{tabular}{c|cc|c|cccc|c|c}
    \toprule
    Dropout & \multicolumn{3}{c|}{In-domain} & \multicolumn{5}{c|}{Out-of-domain} & \multirow{2}{*}{Average}\\
    Probability & High-Quality Test Set & Los Angeles & Average & Mid-/Low-Quality Test Set & Sao Paulo & Medellin & Guangzhou & Average & \\ \midrule
        0.3 & 4.2323 & \firstbest{3.5862} & \firstbest{3.9093} & 9.2725 & 5.5070 & \firstbest{3.8394} & \firstbest{12.5761} & 7.7987 & \firstbest{6.5023} \\ \midrule
        0.1 & 4.5570 & 6.5463 & 5.5517 & \firstbest{8.6084} & 5.4523 & 3.8491 & 13.0941 & \firstbest{7.7510} & 7.0179 \\
        0.5 & 4.1486 & 3.8777 & 4.0131 & 8.9927 & 5.7659 & 4.9133 & 13.4232 & 8.2738 & 6.8536 \\ 
        0.7 & \firstbest{3.9713} & 3.9574 & 3.9644 & 9.9563 & \firstbest{5.4374} & 4.1067 & 12.6887 & 8.0473 & 6.6863 \\ 
        \bottomrule
    \end{tabular}}
    \label{tab:do_gbh}
\end{table*}
\section{Conclusions}\label{chap:conclusions}
We propose a novel pipeline to enhance monocular height estimation networks by effectively leveraging imperfect labels through an ensemble architecture with specialized branches tailored to different label-quality levels. Weak supervision from mid- and low-quality labels is incorporated via balanced soft height loss and ordinal relation constraints. Further, prediction sharpness is improved using ground truth augmentation. Extensive experiments demonstrate that our pipeline successfully utilizes imperfect labels, enhancing performance across diverse domains.

This approach is particularly beneficial for regions with limited high-quality annotations, making it well-suited for large-scale applications. Nonetheless, it does not address cases with entirely unlabeled data. Future work may focus on extending the framework to incorporate unlabeled samples, thereby further enhancing its generalization capabilities. Additionally, this study primarily targets building height estimation. Its applicability to other ground objects, such as vegetation, remains to be explored due to limited data availability.
\section*{Acknowledgments}
The authors would like to thank the IEEE GRSS Image Analysis and Data Fusion Technical Committee, Aerospace Information Research Institute, Chinese Academy of Sciences, Universität der Bundeswehr München, and GEOVIS Earth Technology Co., Ltd. for organizing the Data Fusion Contest.
\section*{Data Availability}
Code is available at \url{https://github.com/zhu-xlab/weakim2h}.
\bibliographystyle{elsarticle-harv} 
\bibliography{JAG2025}

\end{document}